\documentclass[runningheads]{llncs}
\usepackage{graphicx}

\usepackage{tikz}
\usepackage{comment} 
\usepackage{amsmath,amssymb} 
\usepackage{color}
\usepackage{appendix}

\usepackage{mathrsfs}
\usepackage{xcolor}
\usepackage{textcomp}
\usepackage{color}
\usepackage{colortbl}
\usepackage{subcaption}
\usepackage{paralist}
\usepackage[nice]{nicefrac}
\usepackage{listings}
\usepackage{nicefrac}
\usepackage{textcomp}  
\usepackage{xfrac}
\usepackage{esvect}

\definecolor{yellow}{rgb}{1,1, 0.6}
\definecolor{orange}{rgb}{1, 0.8, 0.6}
\definecolor{red}{rgb}{1, 0.6, 0.6}
\definecolor{darkred}{rgb}{0.8, 0.0, 0.0}
\definecolor{blue}{rgb}{0, 0, 1.0}
\definecolor{green}{rgb}{0, 1.0, 0}
\definecolor{darkgreen}{rgb}{0, 0.4, 0}
\definecolor{darkorange}{rgb}{1, 0.6, 0}
\definecolor{lightgray}{rgb}{0.9, 0.9, 0.9}

\newcommand{\norm}[1]{\left\lVert#1\right\rVert}

\newcommand{\cumsum}[1]{\operatorname{csum}\!\left(#1\right)}
\newcommand{\decsum}[1]{\operatorname{dsum}\!\left(#1\right)}
\newcommand{\argmax}[1]{\operatorname{arg\,max}_i\left(#1\right)}
\newcommand{\argmin}[1]{\operatorname{arg\,min}_i\left(#1\right)}
\newcommand{\myvec}[1]{\boldsymbol{#1}}

\newcommand{\otsuscorescalar}{o}
\newcommand{\otsuscore}{\myvec{\otsuscorescalar}}
\newcommand{\threshold}{t}
\newcommand{\myimage}{\mathrm{I}}
\newcommand{\xvec}{\myvec{x}}
\newcommand{\nvec}{\myvec{n}}

\newcommand{\intweightscalar}{w}
\newcommand{\intweight}{\myvec{\intweightscalar}}

\newcommand{\distortion}{\myvec{d}}

\newcommand{\stdscalar}{\sigma}

\newcommand{\varscalar}{v}
\newcommand{\var}{\myvec{\varscalar}}

\newcommand{\meanscalar}{\mu}
\newcommand{\mean}{\myvec{\meanscalar}}

\newcommand{\mixturescalar}{\pi}
\newcommand{\mixture}{\myvec{\mixturescalar}}

\newcommand{\myscorescalar}{f}
\newcommand{\myscore}{\myvec{\myscorescalar}}

\newcommand{\metscorescalar}{\ell}
\newcommand{\metscore}{\myvec{\metscorescalar}}

\newcommand{\sumvec}[1]{\norm{#1}_1}
\newcommand{\nsum}{\sumvec{\nvec}}

\newcommand{\Xnu}{\nu}
\newcommand{\regstrength}{\kappa}
\newcommand{\Xsigma}{\tau}
\newcommand{\regtarget}{\omega}

\newcommand{\nveclen}{\operatorname{len}(\nvec)\mbox{-}1}

\newcommand{\sichisq}{\chi^2_{\mathrm{SI}}} 
\newcommand{\mymid}{{\,\mid\,}}

\newcommand{\etal}{\textit{et al}. }
\newcommand{\etc}{\textit{etc}.}
\newcommand{\ie}{\textit{i}.\textit{e}., }
\newcommand{\eg}{\textit{e}.\textit{g}. }

\newcommand{\at}[2]{#1_{#2}}

\newcommand{\blursigma}{\sigma}
\newcommand{\blurvector}{\boldsymbol{f}(\blursigma)}

\newcommand{\dotprod}[2]{\left\langle #1, #2 \right\rangle}

\usepackage[width=122mm,left=12mm,paperwidth=146mm,height=193mm,top=12mm,paperheight=217mm]{geometry}

\begin{document}
\pagestyle{headings}
\mainmatter
\def\ECCVSubNumber{3657}  

\title{A Generalization of Otsu's Method and Minimum Error Thresholding}

\titlerunning{A Generalization of Otsu's Method and Minimum Error Thresholding}
%
\author{Jonathan T. Barron}
\authorrunning{J. T. Barron}
%
\institute{Google Research \\
\email{barron@google.com}}
\maketitle

\begin{abstract}

We present Generalized Histogram Thresholding (GHT), a simple, fast, and effective technique for histogram-based image thresholding.
GHT works by performing approximate maximum a posteriori estimation of a mixture of Gaussians with appropriate priors.
We demonstrate that GHT subsumes three classic thresholding techniques as special cases: Otsu's method, Minimum Error Thresholding (MET), and weighted percentile thresholding.
GHT thereby enables the continuous interpolation between those three algorithms, which allows thresholding accuracy to be improved significantly.
GHT also provides a clarifying interpretation of the common practice of coarsening a histogram's bin width during thresholding.
We show that GHT outperforms or matches the performance of all algorithms on a recent challenge for handwritten document image binarization (including deep neural networks trained to produce per-pixel binarizations),
and can be implemented in a dozen lines of code or as a trivial modification to Otsu's method or MET.
\end{abstract}

\begin{figure}[t]
    \centering 
    \begin{subfigure}[b]{0.49\linewidth}
        \centering
        \frame{\includegraphics[width=\linewidth]{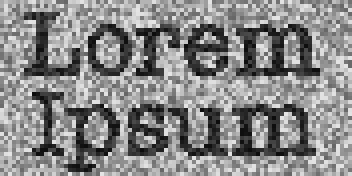}}
        \caption{An input image $\myimage$}
        \label{fig:teaser1}
    \end{subfigure}
    \begin{subfigure}[b]{0.48\linewidth}
        \centering
        \includegraphics[width=\linewidth]{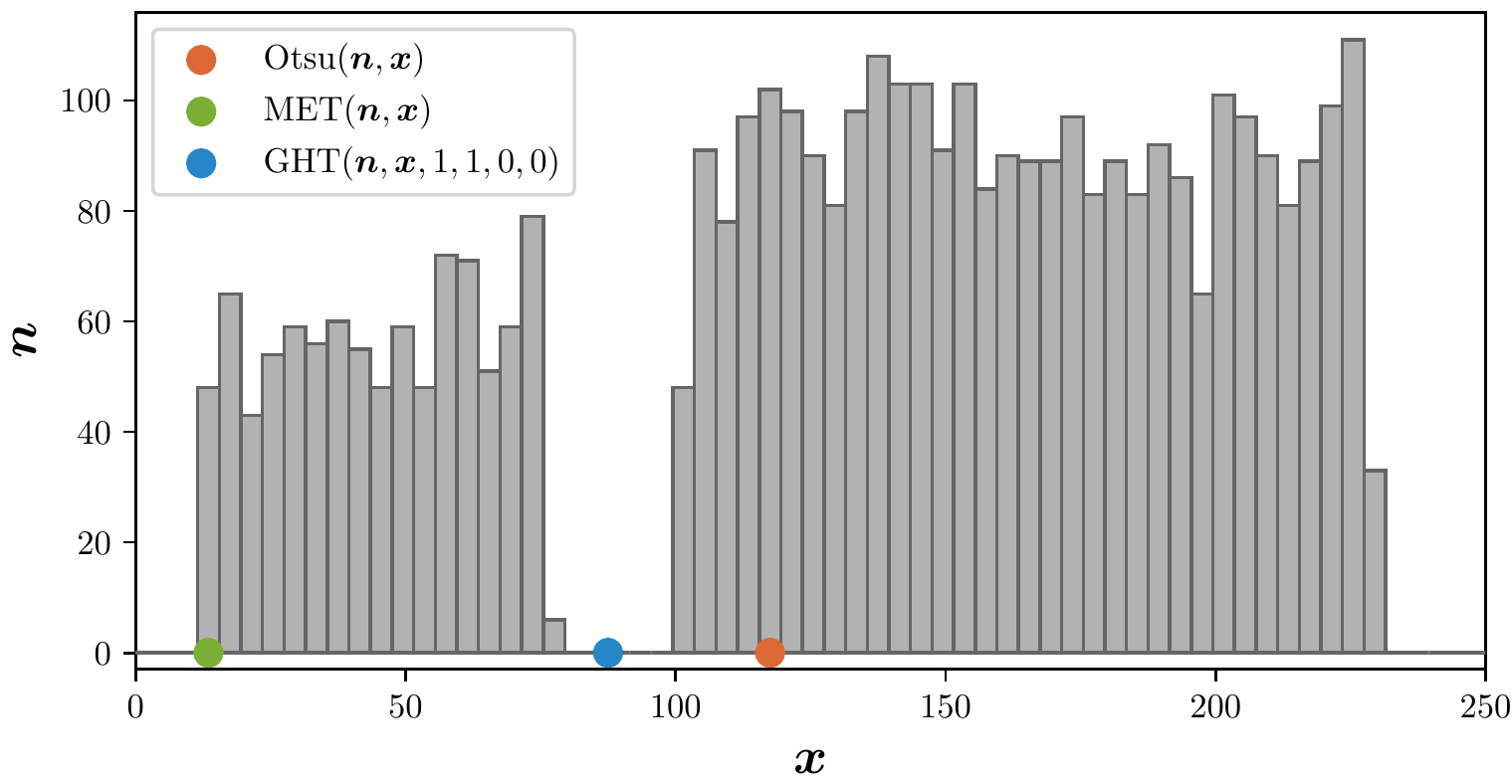}
        \caption{$(\nvec, \xvec) = \operatorname{hist}(\myimage)$}
        \label{fig:teaser2}
    \end{subfigure}
    \begin{tabular}{@{}c@{\quad}c@{\quad}c@{}}
    \begin{subfigure}[b]{0.31\linewidth}
        \centering
        \frame{\includegraphics[width=\linewidth]{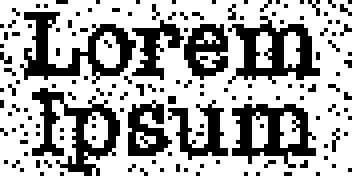}}
        \caption{$\myimage > \operatorname{Otsu}(\nvec, \xvec)$}
        \label{fig:teaser3}
    \end{subfigure}&
    \begin{subfigure}[b]{0.31\linewidth}
        \centering
        \frame{\includegraphics[width=\linewidth]{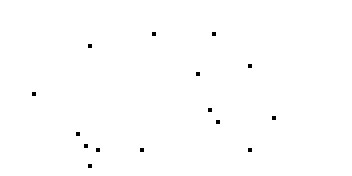}}
        \caption{$\myimage > \operatorname{MET}(\nvec, \xvec)$}
        \label{fig:teaser4}
    \end{subfigure}&
    \begin{subfigure}[b]{0.31\linewidth}
        \centering
        \frame{\includegraphics[width=\linewidth]{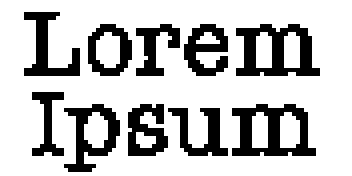}}
        \caption{$\myimage > \operatorname{GHT}(\nvec, \xvec; 1, 1, 0, \cdot)$}
        \label{fig:teaser5}
    \end{subfigure}
    \end{tabular}
    \caption{
    Despite the visually apparent difference between foreground and background pixels in the image (a) and its histogram (b), both Otsu's method (c) and MET (d) fail to correctly binarize this image.
    GHT (e), which includes both Otsu's method and MET as special cases, produces the expected binarization.
    }
    \label{fig:teaser}
\end{figure}

\section{Introduction}

Histogram-based thresholding is a ubiquitous tool in image processing, medical imaging, and document analysis: The grayscale intensities of an input image are used to compute a histogram, and some algorithm is then applied to that histogram to identify an optimal threshold (corresponding to a bin location along the histogram's $x$-axis) with which the histogram is to be ``split'' into two parts. That threshold is then used to binarize the input image, under the assumption that this binarized image will then be used for some downstream task such as classification or segmentation.
Thresholding has been the focus of a half-century of research, and is well documented in several survey papers \cite{sahoo1988survey,zhang2008image}.
One might assume that a global thresholding approach has little value in a modern machine learning context: preprocessing an image via binarization discards potentially-valuable information in the input image, and training a neural network to perform binarization is a straightforward alternative.
However, training requires significant amounts of training data, and in many contexts (particularly medical imaging) such data may be prohibitively expensive or difficult to obtain. As such, there continues to be value in ``automatic'' thresholding algorithms that do not require training data, as is evidenced by the active use of these algorithms in the medical imaging literature.

\section{Preliminaries}

Given an input image $\myimage$ (Figure~\ref{fig:teaser1}) that is usually assumed to contain 8-bit or 16-bit quantized grayscale intensity values, a histogram-based automatic thresholding technique first constructs a histogram of that image consisting of a vector of histogram bin locations $\xvec$ and a corresponding vector of histogram counts or weights $\nvec$  (Figure~\ref{fig:teaser2}).
With this histogram, the algorithm attempts to find some threshold $\threshold$ that separates the two halves of the histogram according to some criteria (\eg maximum likelihood, minimal distortion, \etc), and this threshold is used to produce a binarized image $\myimage > \threshold$ (Figures~\ref{fig:teaser3}-\ref{fig:teaser5}).
Many prior works assume that $\at{x}{i} = i$, which addresses the common case in which histogram bins exactly correspond to quantized pixel intensities, but we will make no such assumption.
This allows our algorithm (and our descriptions and implementations of baseline algorithms) to be applied to arbitrary sets of sorted points, and to histograms whose bins have arbitrary locations. For example, it is equivalent to binarize some quantized image $\myimage$ using a histogram:
\begin{equation}
(\nvec, \xvec) = \operatorname{hist}(\myimage)\,,
\end{equation}
or using a vector of sorted values each with a ``count'' of 1:
\begin{equation}
 \nvec = \vv{1}\,, \quad\quad \xvec = \operatorname{sort}(\myimage)\,.
\label{eq:sorted}
\end{equation}
This equivalence is not used in any result presented here, but is useful when binarizing a set of continuous values.

Most histogram-based thresholding techniques work by considering each possible ``split'' of the histogram: each value in $\xvec$ is considered as a candidate for $\threshold$, and two quantities are produced that reflect the surface statistics of $\at{\nvec}{\xvec{\,\leq\,}\threshold}$ and $\at{\nvec}{\xvec{\,>\,}\threshold}$.
A critical insight of many classic histogram-based thresholding techniques is that some of these quantities can be computed recursively. 
For example, the sum of all histogram counts up through some index $i$ ($\at{\intweightscalar^{(0)}}{i} = \sum_{j=0}^i \at{n}{j}$) need not be recomputed from scratch if that sum has already been previously computed for all the previous histogram element, and can instead just be updated with a single addition ($\at{\intweightscalar^{(0)}}{i} = \at{\intweightscalar^{(0)}}{i\,\mbox{--}1} + \at{n}{i}$).
Here we construct pairs of vectors of intermediate values that measure surface statistics of the histogram above and below each ``split'', to be used by GHT and our baselines:
\begin{align}
\intweight^{(0)} &= \cumsum{\nvec} &
\intweight^{(1)} &= \decsum{\nvec} \label{eq:prelim} \\
\mixture^{(0)} &= \intweight^{(0)} / \nsum &
\mixture^{(1)} &= \intweight^{(1)} / \nsum = 1-\mixture^{(0)} \nonumber \\
\mean^{(0)} &= \cumsum{\nvec \xvec} / \intweight^{(0)} \nonumber & 
\mean^{(1)} &= \decsum{\nvec \xvec} / \intweight^{(1)} \nonumber\\
\distortion^{(0)} &= \cumsum{\nvec \xvec^2} - \intweight^{(0)} \left(\mean^{(0)}\right)^2 \nonumber &
\distortion^{(1)} &= \decsum{\nvec \xvec^2} - \intweight^{(1)} \left(\mean^{(1)}\right)^2 \nonumber
\end{align}
All multiplications, divisions, and exponentiations are element-wise, and $\cumsum{}$ and $\decsum{}$ are cumulative and ``reverse cumulative'' sums respectively.
$\nsum$ is the sum of all histogram counts in $\nvec$, while $\intweight^{(0)}$ and $\intweight^{(1)}$ are the sums of histogram counts in $\nvec$ below and above each split of the histogram, respectively.
Similarly, each $\mixture^{(k)}$ respectively represent normalized histogram counts (or mixture weights) above and below each split.
Each $\mean^{(k)}$ and $\distortion^{(k)}$ represent the mean and distortion of all elements of $\xvec$ below and above each split, weighted by their histogram counts $\nvec$.
Here ``distortion'' is used in the context of k-means, where it refers to the sum of squared distances of a set of points to their mean.
The computation of $\distortion^{(k)}$ follows from three observations:
First, $\cumsum{\cdot}$ or $\decsum{\cdot}$ can be used (by weighting by $\nvec$ and normalizing by $\intweight$) to compute a vector of expected values. Second, the sample variance of a quantity is a function of the expectation of its values and its values squared ($\operatorname {Var} (X) = \operatorname {E}[X^{2}]-\operatorname {E} [X]^{2}$). Third, the sample variance of a set of points is proportional to the total distortion of those points.

Formally, $\at{\cumsum{\cdot}}{i}$ is the inclusive sum of every vector element at or before index $i$, and $\at{\decsum{\cdot}}{i}$ is the exclusive sum of everything after index $i$:
\begin{align}
    \at{\cumsum{\nvec}}{i} &= \sum_{j=0}^{i} \at{n}{j} = \at{\cumsum{\nvec}}{i-1} + \at{n}{i}\,, \\
    \at{\decsum{\nvec}}{i} &= \!\!\!\!\!\sum_{j=i+1}^{\nveclen}\!\!\!\! \at{n}{j} = \at{\decsum{\nvec}}{i+1} + \at{n}{i+1}\,. \nonumber 
\end{align}
Note that the outputs of these cumulative and decremental sums have a length that is one less than that of $\xvec$ and $\nvec$, as these values measure ``splits'' of the data, and we only consider splits that contain at least one histogram element.
The subscripts in our vectors of statistics correspond to a threshold \emph{after} each index, which means that binarization should be performed by a greater-than comparison with the returned threshold ($\xvec > \threshold$).
In practice, $\decsum{\nvec}$ can be computed using just the cumulative and total sums of $\nvec$:
\begin{align}
    \decsum{\nvec} &= \nsum - \cumsum{\nvec}\,.
\end{align}
As such, these quantities (and all other vector quantities that will be described later) can be computed efficiently with just one or two passes over the histogram.

\section{Algorithm}

Our Generalized Histogram Thresholding (GHT) algorithm is motivated by a straightforward Bayesian treatment of histogram thresholding.
We assume that each pixel in the input image is generated from a mixture of two probability distributions corresponding to intensities below and above some threshold, and we maximize the joint probability of all pixels (which are assumed to be i.i.d.):
\begin{gather}
    \prod_{x,y} \sum_{k=1}^2 p\Big(\myimage_{x,y} \, \Big| \, z_{x,y}=k \Big) p(z_{x,y}=k) \,.
    \label{eq:bayesianprob}
\end{gather}
The probability of each pixel intensity $\myimage_{x,y}$ conditioned on its (unknown) mixture component $z_{x,y}$ is:
\begin{equation}
\resizebox{\textwidth}{!}{$
    \displaystyle
    p\Big(\myimage_{x,y} \, \Big| \, z_{x,y}=k \Big) = \mixturescalar^{(k)} \mathcal{N}\!\left( \myimage_{x,y}  \mymid  \meanscalar^{(k)}, \stdscalar^{(k)} \right) 
    \sichisq\!\left(\stdscalar^{(k)}  \mymid \mixturescalar^{(k)} \Xnu, \Xsigma^2 \right)
     \operatorname{Beta}\!\left( \mixturescalar^{(k)}  \mymid  \regstrength, \regtarget \right)\,.
     \label{eq:probform}
     $}
\end{equation}
This is similar to a straightforward mixture of Gaussians: we have a mixture probability ($\mixturescalar^{(0)}$, $\mixturescalar^{(1)} = 1 - \mixturescalar^{(0)}$), two means ($\meanscalar^{(0)}$, $\meanscalar^{(1)}$), and two standard deviations ($\stdscalar^{(0)}$, $\stdscalar^{(1)}$).
But unlike a standard mixture of Gaussians, we place conjugate priors on model parameters $\mixturescalar^{(k)}$ and $\stdscalar^{(k)}$. We assume each standard deviation is the mode (\ie the ``updated'' $\Xsigma$ value) of a scaled inverse chi-squared distribution, which is the conjugate prior of a Gaussian with a known mean.
\begin{equation}
\sichisq(x \mymid \Xnu, \Xsigma^2) = {\frac {(\Xsigma ^{2}\Xnu /2)^{\Xnu /2}}{\Gamma (\Xnu /2)}}~{\frac {\exp\!\left({\frac {-\Xnu \Xsigma ^{2}}{2 x}}\right)}{x^{1+\Xnu /2}}}\,.
\end{equation}
And we assume the mixture probability is drawn from a beta distribution (the conjugate prior of a Bernoulli distribution).
\begin{equation}
\operatorname{Beta}(x \mymid \regstrength, \regtarget) = {\frac {x^{\alpha -1}(1-x)^{\beta -1}}{\operatorname{B} (\alpha ,\beta )}}\,, \quad\,\,
  \alpha = \regstrength \regtarget + 1\,, \quad\,\,
  \beta = \regstrength(1 - \regtarget) + 1\,.
\end{equation}
Where $\operatorname{B}$ is the beta function.
Note that our beta distribution uses a slightly unconventional parameterization in terms of its concentration $\regstrength$ and mode $\regtarget$, which will be explained later. Also note that in Equation~\ref{eq:probform} the degrees-of-freedom parameter $\Xnu$ of the scaled inverse chi-squared distribution is rescaled according to the mixture probability, as will also be discussed later.
This Bayesian mixture of Gaussians has four hyperparameters that will determine the behavior of GHT: $\Xnu \geq 0, \Xsigma \geq 0, \regstrength \geq 0, \regtarget \in [0, 1]$.

From this probabilistic framework we can derive our histogram thresholding algorithm.
Let us rewrite Equation~\ref{eq:bayesianprob} in terms of histogram locations $\xvec$ and histogram counts $\nvec$ with the summation written explicitly, and then take its logarithm to produce a log-likelihood:
\begin{equation}
    \sum_i \at{n}{i} \log\!\bigg( p\Big( \at{x}{i} \, \Big| \, z_i=0 \Big) p(z_i=0)\,+\, p\Big( \at{x}{i} \, \Big| \, z_i=1 \Big) p(z_i=1) \bigg)\,.
    \label{eq:bayesianloghist}
\end{equation}
Now we make a simplifying model assumption that enables a single-pass histogram thresholding algorithm.
We will assume that, at each potential ``split'' of the data, each of the two splits of the histogram is generated entirely by one of the two Gaussians in our mixture.
More formally, we consider all possible sorted assignments of each histogram bin to either of the two Gaussians and maximize the expected complete log-likelihood (ECLL) of that assignment.
Jensen's inequality ensures that this ECLL is a lower bound on the marginal likelihood of the data that we would actually like to maximize.
This is the same basic approach (though not the same justification) as in ``Minimum Error Thresholding'' (MET)~\cite{KittlerIllingworth} and other similar approaches, but where our technique differs is in how the likelihood and the parameters of these Gaussians are estimated.
For each split of the data, we assume the posterior distribution of the latent variables determining the ownership of each histogram bin by each Gaussian (the unknown $z$ values) are wholly assigned to one of the two Gaussians according to the split.
This gives us the following ECLL as a function of the assumed split location's array index $i$:
\newcommand{\bhtecll}{\mathcal{L}_i}
\begin{equation}
    \bhtecll = \sum_{j=0}^i \at{n}{j} \log\!\bigg( p\Big( \at{x}{j} \, \Big| \, z_j=0 \Big) \bigg) + \sum_{j=i+1}^{\nveclen} \at{n}{j} \log\!\bigg(  p\Big( \at{x}{j} \, \Big| \, z_j=1 \Big) \bigg)\,.
    \label{eq:ecll}
\end{equation}
Our proposed algorithm is to simply iterate over all possible values of $i$ (of which there are $255$ in our experiments) and return the value that maximizes $\bhtecll$.
As such, our algorithm can be viewed as a kind of inverted expectation-maximization in which the sweep over threshold resembles an M-step and the assignment of latent variables according to that threshold resembles an E-step \cite{SalakhutdinovRG03}.
Our GHT algorithm will be defined as:
\begin{equation}
    \operatorname{GHT}(\xvec,  \nvec; \Xnu, \Xsigma, \regstrength, \regtarget) = \at{x}{\argmax{\bhtecll}} \,. \label{eq:threshprob}
\end{equation}
This definition of GHT is a bit unwieldy, but using the preliminary math of 
Equation~\ref{eq:prelim} and ignoring global shifts and scales of $\bhtecll$ that do not affect the argmax it can be simplified substantially:
\begin{align}
  \var^{(0)} &= \frac{\mixture^{(0)} \Xnu \Xsigma^2 + \distortion^{(0)}}{\mixture^{(0)} \Xnu + \intweight^{(0)}} \quad\quad\quad
  \var^{(1)} = \frac{\mixture^{(1)} \Xnu \Xsigma^2 + \distortion^{(1)}}{\mixture^{(1)} \Xnu + \intweight^{(1)}} \nonumber \\
  \myscore^{(0)} &= -\frac{\distortion^{(0)}}{\var^{(0)}} - \intweight^{(0)} \log\!\left(\!\var^{(0)}\!\right)  + 2\!\left( \intweight^{(0)}\!+\!\regstrength\regtarget \right) \log\!\left(\!\intweight^{(0)}\!\right)  \nonumber \\
  \myscore^{(1)} &= -\frac{\distortion^{(1)}}{\var^{(1)}}  - \intweight^{(1)} \log\!\left(\!\var^{(1)}\!\right)  + 2\!\left( \intweight^{(1)}\!+\!\regstrength(1\!-\!\regtarget) \right) \log\!\left(\!\intweight^{(1)}\!\right) \nonumber  \\
  & \operatorname{GHT}(\xvec,  \nvec; \Xnu, \Xsigma, \regstrength, \regtarget) = \at{x}{\argmax{\myscore^{(0)} + \myscore^{(1)}}} \label{eq:thresh}
\end{align}
Each $\myscore^{(k)}$ can be thought of as a log-likelihood of the data for each Gaussian at each split, and each $\var^{(k)}$ can be thought of as an estimate of the variance of each Gaussian at each split.
GHT simply scores each split of the histogram and returns the value of $\xvec$ that maximizes the $\myscore^{(0)} + \myscore^{(1)}$ score (in the case of ties we return the mean of all $\xvec$ values that maximize that score).
Because this only requires element-wise computation using the previously-described quantities in Equation~\ref{eq:prelim}, GHT requires just a single linear scan over the histogram.

The definition of each $\var^{(k)}$ follows from our choice to model each Gaussian's variance with a scaled inverse chi-squared distribution: the $\Xnu$ and $\Xsigma^2$ hyperparameters of the scaled inverse chi-squared distribution are updated according to the scaled sample variance (shown here as distortion $\distortion^{(k)}$) to produce an updated posterior hyperparameter that we use as the Gaussian's estimated variance  $\var^{(k)}$.
Using a conjugate prior update instead of the sample variance has little effect when the subset of $\nvec$ being processed has significant mass (\ie when $\intweight^{(k)}$ is large) but has a notable effect when a Gaussian is being fit to a sparsely populated subset of the histogram.
Note that $\var^{(k)}$ slightly deviates from the traditional conjugate prior update, which would omit the $\mixture^{(k)}$ scaling on $\Xnu$ in the numerator and denominator (according to our decision to rescale each degrees-of-freedom parameter by each mixture probability).
This additional scaling counteracts the fact that, at different splits of our histogram, the total histogram count on either side of the split will vary. A ``correct'' update would assume a constant number of degrees-of-freedom regardless of where the split is being performed, which would result in the conjugate prior here having a substantially different effect at each split, thereby making the resulting $\myscore^{(0)} + \myscore^{(1)}$ scores not comparable to each other and making the argmax over these scores not meaningful.

The beta distribution used in our probabilistic formulation (parameterized by a concentration $\regstrength$ and a mode $\regtarget$) has a similar effect as the ``anisotropy coefficients'' used by other models \cite{kapur1985new,pun1981entropic}: setting $\regtarget$ near $0$ biases the algorithm towards thresholding near the start of the histogram, and setting it near $1$ biases the algorithm towards thresholding near the end of the histogram. The concentration $\regstrength$ parameter determines the strength of this effect, and setting $\regstrength=0$ disables this regularizer entirely.
Note that our parameterization of concentration $\regstrength$ differs slightly from the traditional formulation, where setting $\regstrength=1$ has no effect and setting $\regstrength < 1$ moves the predicted threshold \emph{away} from the mode. We instead assume the practitioner will only want to consider biasing the algorithm's threshold \emph{towards} the mode, and parameterize the distribution accordingly.
These hyperparameters allow for GHT to be biased towards outputs where a particular fraction of the image lies above or below the output threshold. For example, in an OCR / digit recognition context, it may be useful to inform the algorithm that the majority of the image is expected to not be ink, so as to prevent output thresholds that erroneously separate the non-ink background of the page into two halves.

Note that, unlike many histogram based thresholding algorithms, we do not require or assume that our histogram counts $\nvec$ are normalized. In all experiments, we use raw counts as our histogram values. This is important to our approach, as it means that the behavior of the magnitude of $\nvec$ and the behavior induced by varying our $\Xnu$ and $\regstrength$ hyperparameters is consistent:
\newcommand{\mult}{a}
\begin{equation}
\forall \mult\in\mathbb{R}_{>0}\,\, \operatorname{GHT}(\xvec, \mult \nvec; \mult \Xnu, \Xsigma, \mult \regstrength, \regtarget) = \operatorname{GHT}(\nvec, \xvec; \Xnu, \Xsigma, \regstrength, \regtarget) \,.
\end{equation}
This means that, for example, doubling the number of pixels in an image is equivalent to halving the values of the two hyperparameters that control the ``strength'' of our two Bayesian model components.

Additionally, GHT's output and hyperparameters (excluding $\Xsigma$, which exists solely to encode absolute scale) are invariant to positive affine transformations of the histogram bin centers:
\newcommand{\shift}{b}
\begin{equation}
\forall \mult\in\mathbb{R}_{>0}\,\, \forall \shift\in\mathbb{R}\,\, \operatorname{GHT}(\mult \xvec + \shift, \nvec; \Xnu, \mult \Xsigma, \regstrength, \regtarget) = \operatorname{GHT}(\nvec, \xvec; \Xnu, \Xsigma, \regstrength, \regtarget) \,.
\end{equation}
One could extend GHT to be sensitive to absolute intensity values by including conjugate priors over the means of the Gaussians, but we found little experimental value in that level of control. This is likely because the dataset we evaluate on does not have a standardized notion of brightness or exposure (as is often the case for binarization tasks).

In the following subsections we demonstrate how GHT generalizes three standard approaches to histogram thresholding by including them as special cases of our hyperparameter settings: Minimum Error Thresholding~\cite{KittlerIllingworth}, Otsu's method~\cite{Otsu}, and weighted percentile thresholding~\cite{percentile}.

\subsection{Special Case: Minimum Error Thresholding}
\label{subsec:met}

Because Minimum Error Thresholding (MET)~\cite{KittlerIllingworth}, like our approach, works by maximizing the expected complete log-likelihood of the input histogram under a mixture of two Gaussians, it is straightforward to express it using the already-defined quantities in Equation~\ref{eq:prelim}:
\begin{align}
    \metscore = 1 + \intweight^{(0)}\!\log\!\left(\!\frac{\distortion^{(0)}}{\intweight^{(0)}}\!\right) + \intweight^{(1)}\!\log\!\left(\!\frac{\distortion^{(1)}}{\intweight^{(1)}}\!\right) \nonumber - 2\!\left(\!\intweight^{(0)}\!\log\!\left(\!\intweight^{(0)}\!\right) + \intweight^{(1)}\!\log\!\left(\!\intweight^{(1)}\!\right)\!\right)\,, \nonumber \\
    \operatorname{MET}(\xvec,  \nvec) = \at{x}{\argmin{ \metscore}}\,. \span 
\end{align}
Because GHT is simply a Bayesian treatment of this same process, it includes MET as a special case. If we set $\Xnu=0$ and $\regstrength=0$ (under these conditions the values of $\Xsigma$ and $\regtarget$ are irrelevant) in Equation~\ref{eq:thresh}, we see that $\var^{(k)}$ reduces to $\distortion^{(k)} / \intweight^{(k)}$, which causes each score to simplify dramatically:
\begin{equation}
    \Xnu=\regstrength=0 \implies \myscore^{(k)} = -\intweight^{(k)} -\intweight^{(k)} \log\!\left(\!\frac{\distortion^{(k)}}{\intweight^{(k)}}\!\right) + 2 \intweight^{(k)} \log\!\left(\!\intweight^{(k)}\!\right)\,.
\end{equation}
Because $\intweight^{(0)} + \intweight = \nsum$, the score maximized by GHT can be simplified:
\begin{equation}
    \Xnu=\regstrength=0 \implies \myscore^{(0)} + \myscore^{(1)} = 1 - \nsum - \metscore\,.
\end{equation}
We see that, for these particular hyperparameter settings, the total score $\myscore^{(0)} + \myscore^{(1)}$ that GHT maximizes is simply an affine transformation (with a negative scale) of the score  $\metscore$ that MET maximizes. This means that the index that optimizes either quantity is identical, and so the algorithms (under these hyperparameter settings) are equivalent:
\begin{equation}
\operatorname{MET}(\nvec, \xvec) = \operatorname{GHT}(\nvec, \xvec; 0, \cdot, 0, \cdot)\,.
\end{equation}

\subsection{Special Case: Otsu's Method}
\label{subsec:otsu}

Otsu's method for histogram thresholding~\cite{Otsu} works by directly maximizing inter-class variance for the two sides of the split histogram, which is equivalent to indirectly minimizing the total intra-class variance of those two sides~\cite{Otsu}.
Otsu's method can also be expressed using the quantities in Equation~\ref{eq:prelim}:
\begin{equation}
  \otsuscore = \intweight^{(0)} \intweight^{(1)} \left(\mean^{(0)} - \mean^{(1)} \right)^2, \quad \operatorname{Otsu}(\nvec, \xvec) = \at{x}{\argmax{ \otsuscore }}\,.
\end{equation}
To clarify the connection between Otsu's method and GHT, we rewrite the score $\otsuscore$ of Otsu's method as an explicit sum of intra-class variances:
\begin{align}
\otsuscore &= \nsum \dotprod{\nvec}{\xvec^2} - \dotprod{\nvec}{\xvec}^2 - \nsum\!\left(\distortion^{(0)} + \distortion^{(1)}\right)\,. \label{eq:distortion_equiv}
\end{align}
Now let us take the limit of the $\myscore^{(0)} + \myscore^{(1)}$ score that is maximized by GHT as $\Xnu$ approaches infinity:
\begin{equation}
\lim_{\Xnu \rightarrow \infty} \myscore^{(0)} + \myscore^{(1)} = -\frac{\distortion^{(0)} + \distortion^{(1)}}{\Xsigma^2} + 2 \intweight^{(0)} \log\!\left(\!\frac{\intweight^{(0)}}{\Xsigma}\!\right) + 2 \intweight^{(1)} \log\!\left(\!\frac{\intweight^{(1)}}{\Xsigma}\!\right)\,.
\label{eq:otsulim}
\end{equation}
With this we can set the $\Xsigma$ hyperparameter to be infinitesimally close to zero, in which case the score in Equation~\ref{eq:otsulim} becomes dominated by its first term.
Therefore, as $\Xnu$ approaches infinity and $\Xsigma$ approaches zero, the score maximized by GHT is proportional to the (negative) score that is indirectly minimized by Otsu's method:
\begin{equation}
\Xnu \gg 0, \, \Xsigma = \epsilon \implies \myscore^{(0)} + \myscore^{(1)} \approx -\frac{\distortion^{(0)} + \distortion^{(1)}}{\Xsigma^2}\,,
\end{equation}
where $\epsilon$ is a small positive number.
This observation fits naturally with the well-understood equivalence of k-means (which works by minimizing distortion) and the asymptotic behavior of maximum likelihood under a mixture of Gaussians model as the variance of each Gaussian approaches zero~\cite{KulisJ12}.
Given this equivalence between the scores that are optimized by MET and GHT (subject to these specific hyperparameters) we can conclude that Otsu's method is a special case of GHT:
\begin{equation}
\operatorname{Otsu}(\nvec, \xvec) = \lim_{(\Xnu, \Xsigma) \rightarrow (\infty, 0)} \operatorname{GHT}(\nvec, \xvec; \Xnu, \Xsigma, 0, \cdot)\,.
\end{equation}


\begin{figure}[t]
    \centering
    \includegraphics[width=0.85\linewidth]{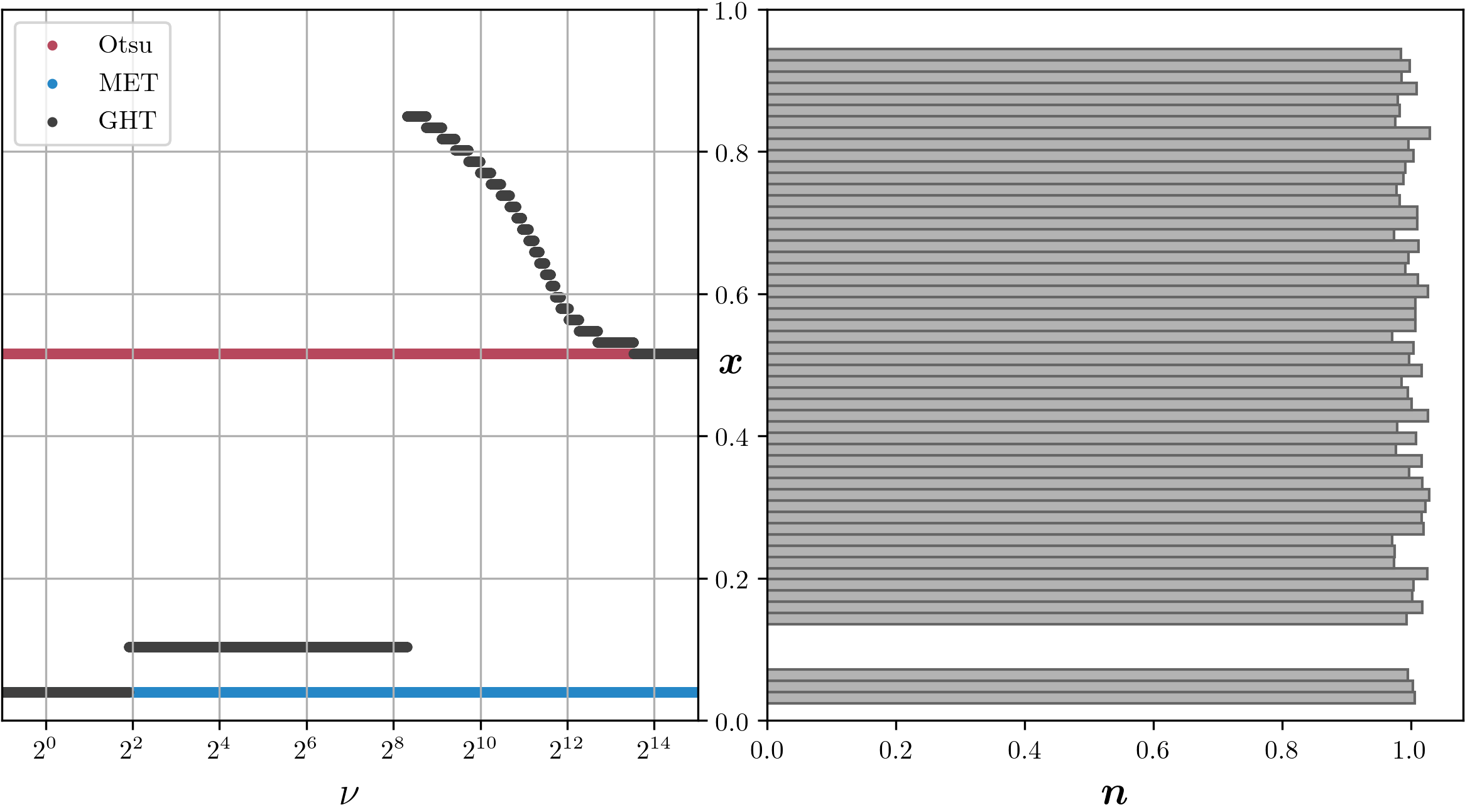}
    \caption{
    On the right we have a toy input histogram (shown rotated), and on the left we have a plot showing the predicted threshold ($y$-axis) of GHT ($\Xsigma=0.01$, $\regstrength=0$, shown in black) relative to the hyperparameter value $\Xnu$ ($x$-axis).
    Note that the $y$-axis is shared across the two plots.
    Both MET and Otsu's method predict incorrect thresholds (shown in red and blue, respectively) despite the evident gap in the input histogram.
    GHT, which includes MET and Otsu's method as special cases ($\Xnu=0$ and $\Xnu=\infty$, respectively) allows us to interpolate between these two extremes by varying $\Xnu$, and produces the correct threshold for a wide range of values $\Xnu \in [\sim\!2^2, \sim\!2^8]$. 
    }
    \label{fig:nu_sweep}
\end{figure}

We have demonstrated that GHT is a generalization of Otsu's method (when $\Xnu = \infty$) and MET (when $\Xnu = 0$), but for this observation to be of any practical importance, there must be value in GHT when $\Xnu$ is set to some value in between those two extremes.
To demonstrate this, in Figure~\ref{fig:nu_sweep} we visualize the effect of $\Xnu$ on GHT's behavior. Similar to the example shown in Figure~\ref{fig:teaser}, we see that both Otsu's method and MET both perform poorly when faced with a simple histogram that contains two separate and uneven uniform distributions: MET prefers to express all histogram elements using a single Gaussian by selecting a threshold at one end of the histogram, and Otsu's method selects a threshold that splits the larger of the two modes in half instead of splitting the two modes apart.
But by varying $\Xnu$ from $0$ to $\infty$ (while also setting $\Xsigma$ to a small value) we see that a wide range of values of $\Xnu$ results in GHT correctly separating the two modes of this histogram. Additionally, we see that the range of hyperparameters that reproduces this behavior is wide, demonstrating the robustness of performance with respect to the specific value of this parameter.

\subsection{Relationship to Histogram Bin Width}

Most histogram thresholding techniques (including GHT as it is used in this paper) construct a histogram with as many bins as there are pixel intensities --- an image of 8-bit integers results in a histogram with 256 bins, each with a bin width of 1. However, the performance of histogram thresholding techniques often depends on the selection of the input histogram's bin width, with a coarse binning resulting in more ``stable'' performance and a fine binning resulting in a more precisely localized threshold. Many histogram thresholding techniques therefore vary the bin width of their histograms, either by treating it as a user-defined hyperparameter~\cite{coudray2010robust} or by determining it automatically~\cite{Kadhim,onumanyi2017modified} using classical statistical techniques~\cite{doane1976aesthetic,sturges1926choice}. Similarly, one could instead construct a fine histogram that is then convolved with a Gaussian before thresholding~\cite{lindblad2000histogram}, which is equivalent to constructing the histogram using a Parzen window. Blurring or coarsening histograms both serve the same purpose of filtering out high frequency variation in the histogram, as coarsening a histogram is equivalent blurring (with a box filter) and then decimating a fine histogram.

This practice of varying histogram bins or blurring a histogram can be viewed through the lens of GHT.
Consider a histogram $(\nvec, \xvec)$, and let us assume (contrary to the equivalence described in Equation~\ref{eq:sorted}) that the spacing between the bins centers in $\xvec$ is constant.
Consider a discrete Gaussian blur filter $\blurvector$ with a standard deviation of $\blursigma$, where that filter is normalized ($\norm{\blurvector}_1=1$).
Let us consider $\nvec * \blurvector$, which is the convolution of histogram counts with our Gaussian blur (assuming a ``full'' convolution of $\nvec$ and $\xvec$ with zero boundary conditions).
This significantly affects the sample variance of the histogram $\varscalar$, which (because convolution is linear) is:
\begin{equation}
\varscalar = \frac{\sum_i \at{(\nvec * \blurvector)}{i} ( \at{x}{i} - \mu)^2}{\nsum} = \frac{\nsum \blursigma^2 + d}{\nsum}\,, 
\quad\quad d = \sum_i \at{n}{i} ( \at{x}{i} - \mu)^2 \,. 
\end{equation}
We use $\varscalar$ and $d$ to describe sample variance and distortion as before, though here these values are scalars as we compute a single estimate of both for the entire histogram.
This sample variance $\varscalar$ resembles the definition of $\var^{(k)}$ in Equation~\ref{eq:thresh}, and we can make the two identical in the limit by setting GHT's hyperparameters to $\Xnu = \epsilon \nsum$ and $\Xsigma = \blursigma / \sqrt{\epsilon}$, where $\epsilon$ is a small positive number.
With this equivalence we see that GHT's $\Xsigma$ hyperparameter serves a similar purpose as coarsening/blurring the input histogram --- blurring the input histogram with a Gaussian filter with standard deviation of $\blursigma$ is roughly equivalent to setting GHT's $\Xsigma$ parameter proportionally to $\blursigma$. Or more formally:
\begin{equation}
\operatorname{MET}\!\left( \nvec, \xvec * \blurvector \right) \approx \operatorname{GHT}\!\left(\nvec, \xvec; \nsum \epsilon, \blursigma / \sqrt{\epsilon}, 0, \cdot \right)\,.
\end{equation}
Unlike the other algorithmic equivalences stated in this work, this equivalence is approximate: changing $\blursigma$ is not exactly equivalent to blurring the input histogram. Instead, it is equivalent to, at each split of $\nvec$, blurring the left and right halves of $\nvec$ independently --- the histogram cannot be blurred ``across'' the split. Still, we observed that these two approaches produce identical threshold results in the vast majority of instances. 

This relationship between GHT's $\Xsigma$ hyperparameter and the common practice of varying a histogram's bin width provides a theoretical grounding of both GHT and other work: GHT may perform well because varying $\Xsigma$ implicitly blurs the input histogram, or prior techniques based on variable bin widths may work well because they are implicitly imposing a conjugate prior on sample variance. This relationship may also shed some light on other prior work exploring the limitations of MET thresholding due to sample variance estimates not being properly ``blurred'' across the splitting threshold~\cite{cho1989improvement}.

\subsection{Special Case: Weighted Percentile}
\label{sec:wprctile}

A simple approach for binarizing some continuous signal is to use the nth percentile of the input data as the threshold, such as its median.
This simple baseline is also expressible as a special case of GHT.
If we set $\regstrength$ to a large value we see that the score $\myscore^{(0)} + \myscore^{(1)}$ being maximized by GHT becomes dominated by the terms of the score related to $\log(\mixture^{(k)})$:
\begin{equation}
  \resizebox{\textwidth}{!}{$
  \displaystyle
  \regstrength \gg 0\!\implies\!\frac{\myscore^{(0)} + \myscore^{(1)}}{2} \approx \left( \nsum + \regstrength \right)\log\left(\nsum\right) + \regstrength \left( \regtarget \log\!\left(\!\mixture^{(0)}\!\right) + (1 - \regtarget) \log\!\left(\!1 - \mixture^{(0)}\!\right) \right)\,.
  $}
\end{equation}
By setting its derivative to zero we see that this score is maximized at the split location $i$ where $\at{\mixturescalar^{(0)}}{i}$ is as close as possible to $\regtarget$. This condition is also satisfied by the $(100\regtarget)$th percentile of the histogrammed data in $\myimage$, or equivalently, by taking the $(100\regtarget)$th weighted percentile of $\xvec$ (where each bin is weighted by $\nvec$).
From this we can conclude that the weighted percentile of the histogram (or equivalently, the percentile of the image) is a special case of GHT:
\begin{equation}
\operatorname{wprctile}(\xvec,  \nvec, 100\regtarget) = \lim_{\regstrength \rightarrow \infty} \operatorname{GHT}(\nvec, \xvec, 0, \cdot, \regstrength, \regtarget)\,.
\end{equation}
This also follows straightforwardly from how GHT uses a beta distribution: if this beta distribution's concentration $\regstrength$ is set to a large value, the mode of the posterior distribution will approach the mode of the prior $\regtarget$.

\begin{figure}[t]
    \centering
    \includegraphics[trim=80 10 70 40, clip, width=0.85\linewidth]{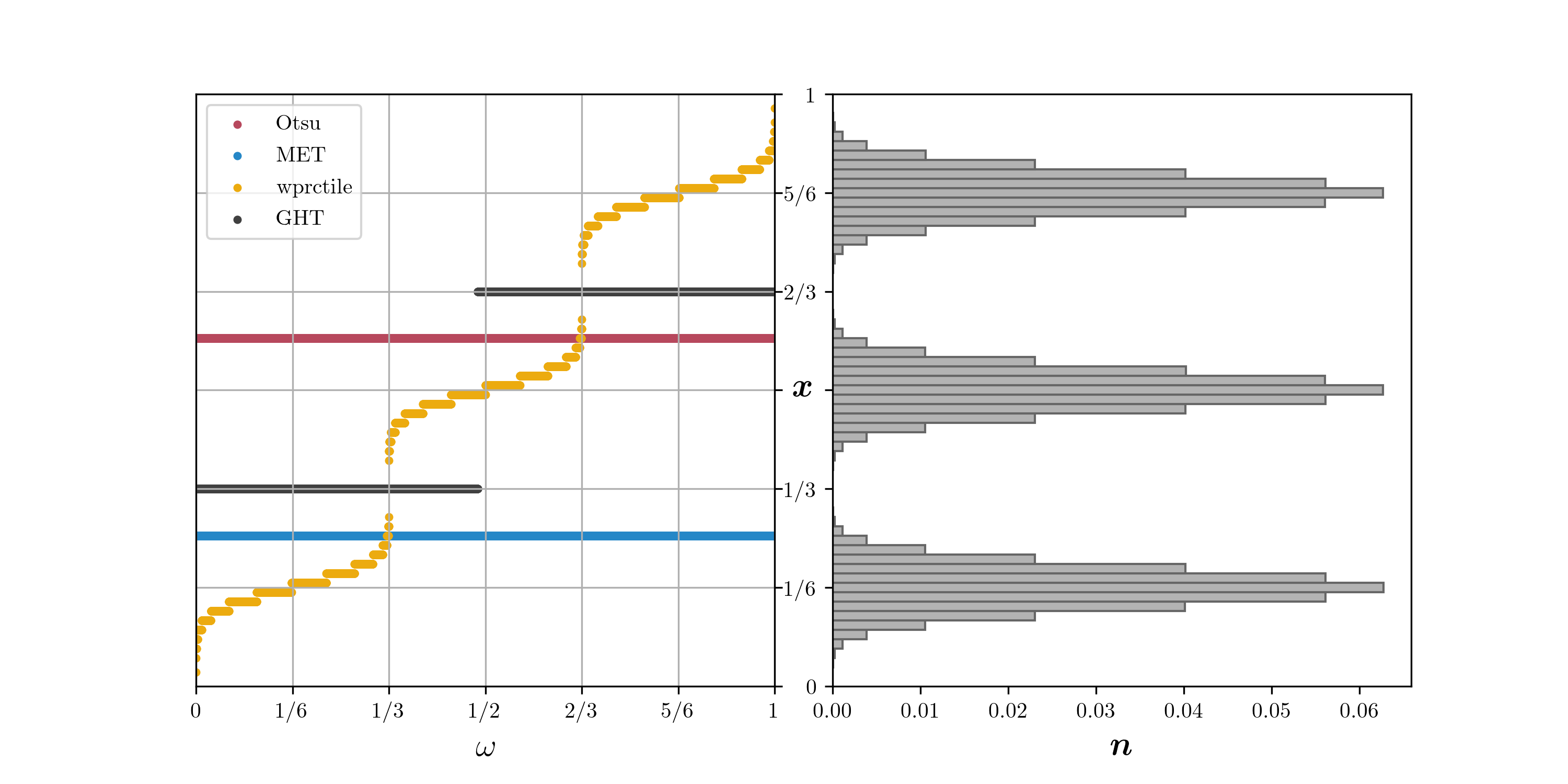}
    \caption{
    On the right we have a toy input histogram (shown rotated), and on the left we have a plot showing the predicted threshold ($y$-axis) of GHT ($\Xnu = 200$, $\Xsigma=0.01$, $\regstrength = 0.1$, shown in black) relative to the hyperparameter value $\regtarget$ ($x$-axis).
    Note that the $y$-axis is shared across the two plots.
    Both MET and Otsu's method (shown in red and blue, respectively) predict thresholds that separate two of the three modes from each other arbitrarily. Computing the weighted percentile of the histogram (shown in yellow) can produce \emph{any} threshold, but reproducing the desired split requires exactly specifying the correct value of $\regtarget$, which likely differs across inputs. GHT (which includes these three other algorithms as special cases)
    produces thresholds that accurately separate the three modes, and the location of that threshold is robust to the precise value of $\regtarget$ and depends only on it being below or above $\sfrac{1}{2}$.
    }
    \label{fig:omega_sweep}
\end{figure}

In Figure~\ref{fig:omega_sweep} we demonstrate the value of this beta distribution regularization.
We see that Otsu's method and MET behave unpredictably when given multi-modal inputs, as the metrics they optimize have no reason to prefer a split that groups more of the histogram modes above the threshold versus one that groups more of the modes below the threshold.
This can be circumvented by using the weighted percentile of the histogram as a threshold, but this requires that the target percentile $\regtarget$ be precisely specified beforehand: For any particular input histogram, it is possible to identify a value for $\regtarget$ that produces the desired threshold, but this percentile target will not generalize to another histogram with the same overall appearance but with a slightly different distribution of mass across that desired threshold level.
GHT, in contrast, is able to produce sensible thresholds for all valid values of $\regtarget$ --- when the value is less than $\sfrac{1}{2}$ the recovered threshold precisely separates the first mode from the latter two, and when the value is greater than $\sfrac{1}{2}$ it precisely separates the first two modes from the latter one. As such, we see that this model component provides an effective means to bias GHT towards specific kinds of thresholds, while still causing it to respect the relative spread of histogram bin counts.


\section{Experiments}

\begin{table*}[h!]
\centering
\resizebox{\textwidth}{!}{  
\begin{tabular}{@{}l|cccc||ccc@{}}
\quad Algorithm & $\Xnu$ & $\Xsigma$ & $\regstrength$ & $\regtarget$ & $F_1 \times 100 \uparrow$ & PSNR $\uparrow$ & DRD~\cite{lu2004distance} $\downarrow$ \\ \hline\hline
{\bf GHT (MET Case)} & -  & -  & -  & -                                                    &                    $60.40 \pm 20.65$ &                    $11.21 \pm 3.50$ &                    $45.32 \pm 41.35$ \\
Kefali \etal \cite{dibco2016,Sari2014TextEF} &&&&                                          &                    $76.10 \pm 13.81$ &                    $15.35 \pm 3.19$ &                    $9.16 \pm 4.87$ \\
Raza \cite{dibco2016} &&&&                                                                 &                    $76.28 \pm 9.71$ &                    $14.21 \pm 2.21$ &                    $15.14 \pm 9.42$ \\
{\bf GHT (wprctile Case)} & -  & -  & $10^{60}$  & $2^{-3.75}$                             &                    $76.77 \pm 14.50$ &                    $15.44 \pm 3.40$ &                    $12.91 \pm 17.19$ \\
Sauvola \cite{dibco2016,Sauvola2000AdaptiveDI} &&&&                                        &                    $82.52 \pm 9.65$ &                    $16.42 \pm 2.87$ &                    $7.49 \pm 3.97$ \\
Khan \& Mollah \cite{dibco2016} &&&&                                                       &                    $84.32 \pm 6.81$ &                    $16.59 \pm 2.99$ &                    $6.94 \pm 3.33$ \\
Tensmeyer \& Martinez \cite{FCN,tensmeyer2017,Wolf2002TextLE} &&&&                         &                    $85.57 \pm 6.75$ &                    $17.50 \pm 3.43$ &                    $5.00 \pm 2.60$ \\
de Almeida \& de Mello \cite{dibco2016} &&&&                                               &                    $86.24 \pm 5.79$ &                    $17.52 \pm 3.42$ &                    $5.25 \pm 2.88$ \\
Otsu's Method \cite{Otsu,dibco2016} &&&&                                                            &                    $86.61 \pm 7.26$ &                    $17.80 \pm 4.51$ &                    $5.56 \pm 4.44$ \\
{\bf GHT (No wprctile)} & $2^{50.5}$  & $2^{0.125}$  & -  & -                              &                    $87.16 \pm 6.32$ &                    $17.97 \pm 4.00$ &                    $5.04 \pm 3.17$ \\
{\bf GHT (Otsu Case)} & $10^{60}$  & $10^{-15}$  & -  & -                                  &                    $87.19 \pm 6.28$ &                    $17.97 \pm 4.01$ &                    $5.04 \pm 3.16$ \\
Otsu's Method (Our Impl.) \cite{Otsu} &&&&                                                 &                    $87.19 \pm 6.28$ &                    $17.97 \pm 4.01$ &                    $5.04 \pm 3.16$ \\
Nafchi \etal - 1 \cite{dibco2016,Nafchi} &&&&                                              &                    $87.60 \pm 4.85$ &                    $17.86 \pm 3.51$ &                    $4.51 \pm 1.62$ \\
Kligler \cite{Howe2013,Katz2007,Kligler2017} &&&&                                          &                    $87.61 \pm 6.99$ &                    $18.11 \pm 4.27$ &                    $5.21 \pm 5.28$ \\
Roe \& de Mello \cite{dibco2016} &&&&                                                      &                    $87.97 \pm 5.17$ &                    $18.00 \pm 3.68$ &                    $4.49 \pm 2.65$ \\
Nafchi \etal - 2 \cite{dibco2016,Nafchi} &&&&                                              &                    $88.11 \pm 4.63$ &                    $18.00 \pm 3.41$ &                    $4.38 \pm 1.65$ \\
Hassa{\"i}ne \etal- 1 \cite{Hassane2011,dibco2016} &&&&                                    &                    $88.22 \pm 4.80$ &                    $18.22 \pm 3.41$ &                    $4.01 \pm 1.49$ \\
Hassa{\"i}ne \etal - 2 \cite{Hassane2012,dibco2016} &&&&                                   & \cellcolor{yellow} $88.47 \pm 4.45$ & \cellcolor{yellow} $18.29 \pm 3.35$ & \cellcolor{orange} $3.93 \pm 1.37$ \\
Hassa{\"i}ne \etal - 3 \cite{Hassane2012,Hassane2011,dibco2016} &&&&                       & \cellcolor{orange} $88.72 \pm 4.68$ & \cellcolor{orange} $18.45 \pm 3.41$ & \cellcolor{red}    $3.86 \pm 1.57$ \\
{\bf GHT} & $2^{29.5}$  & $2^{3.125}$  & $2^{22.25}$  & $2^{-3.25}$                        & \cellcolor{red}    $88.77 \pm 4.99$ & \cellcolor{red}    $18.55 \pm 3.46$ & \cellcolor{yellow} $3.99 \pm 1.77$ \\
\hline
\rowcolor{lightgray} Oracle Global Threshold &&&&                                          & $90.69 \pm 3.92$ & $19.17 \pm 3.29$ & $3.57 \pm 1.84$
\end{tabular}
}
\caption{
Results on the 2016 Handwritten Document Image Binarization Contest (H-DIBCO) challenge \cite{dibco2016}, in terms of the arithmetic mean and standard deviation of F1 score (multiplied by $100$ to match the conventions used by \cite{dibco2016}), PSNR, and Distance-Reciprocal Distortion (DRD)~\cite{lu2004distance}. The scores for all baseline algorithms are taken from \cite{dibco2016}.
Our GHT algorithm and it's ablations and special cases (some of which correspond to other algorithms) are indicated in bold. The settings of the four hyperparameters governing GHT's behavior are also provided.
}
\label{table:results}
\end{table*}

To evaluate GHT, we use the 2016 Handwritten Document Image Binarization Contest (H-DIBCO) challenge \cite{dibco2016}. This challenge consists of images of handwritten documents alongside ground-truth segmentations of those documents, and algorithms are evaluated by how well their binarizations match the ground truth.
We use the 2016 challenge because this is the most recent instantiation of this challenge where a global thresholding algorithm (one that selects a single threshold for use on all pixels) can be competitive:
Later versions of this challenge use input images that contain content outside of the page being binarized which, due to the background of the images often being black, renders any algorithm that produces a single global threshold for binarization ineffective.
GHT's four hyperparameters were tuned using coordinate descent to maximize the arithmetic mean of $F_1$ scores (a metric used by the challenge) over a small training dataset (or perhaps more accurately, a ``tuning'' dataset).
For training data we use the $8$ hand-written images from the 2013 version of the same challenge, which is consistent with the tuning procedure advocated by the challenge (any training data from past challenges may be used, though we found it sufficient to use only relevant data from the most recent challenge).
Input images are processed by taking the per-pixel $\operatorname{max}()$ across color channels to produce a grayscale image, computing a single 256-bin histogram of the resulting values, applying GHT to that histogram to produce a single global threshold, and binarizing the grayscale input image according to that threshold.

\begin{figure}[p]
    \centering 
    \begin{subfigure}[b]{0.325\linewidth}
        \centering
        \includegraphics[width=\linewidth]{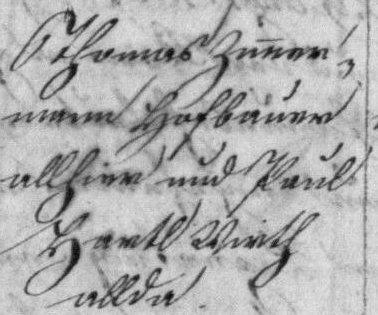}
        \caption{Input Image}

    \end{subfigure}
    \begin{subfigure}[b]{0.325\linewidth}
        \centering
        \frame{\includegraphics[width=\linewidth]{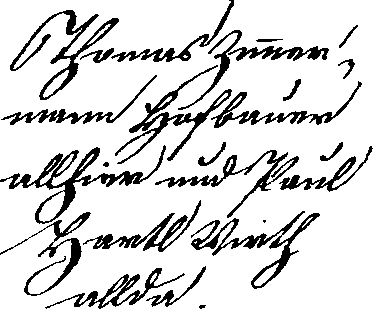}}
        \caption{True Binarization}

    \end{subfigure}
    \begin{subfigure}[b]{0.325\linewidth}
        \centering
        \frame{\includegraphics[width=\linewidth]{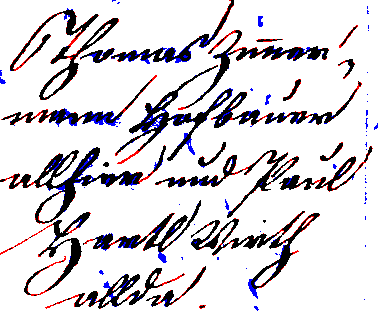}}
        \caption{Oracle Binarization}

    \end{subfigure}
    \begin{subfigure}[b]{0.325\linewidth}
        \centering
        \frame{\includegraphics[width=\linewidth]{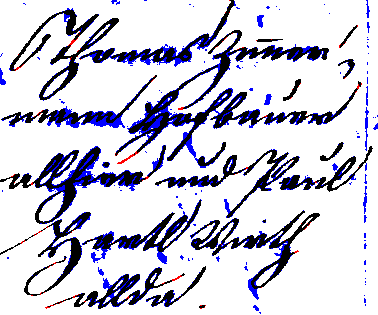}}
        \caption{Otsu's Method}

    \end{subfigure}
    \begin{subfigure}[b]{0.325\linewidth}
        \centering
        \frame{\includegraphics[width=\linewidth]{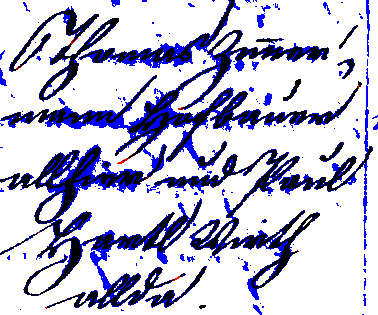}}
        \caption{MET}

    \end{subfigure}
    \begin{subfigure}[b]{0.325\linewidth}
        \centering
        \frame{\includegraphics[width=\linewidth]{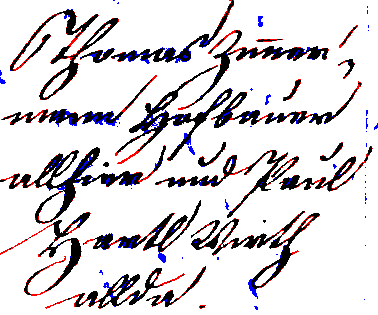}}
        \caption{GHT}

    \end{subfigure}
    \begin{subfigure}[b]{\linewidth}
        \centering
        \includegraphics[width=\linewidth]{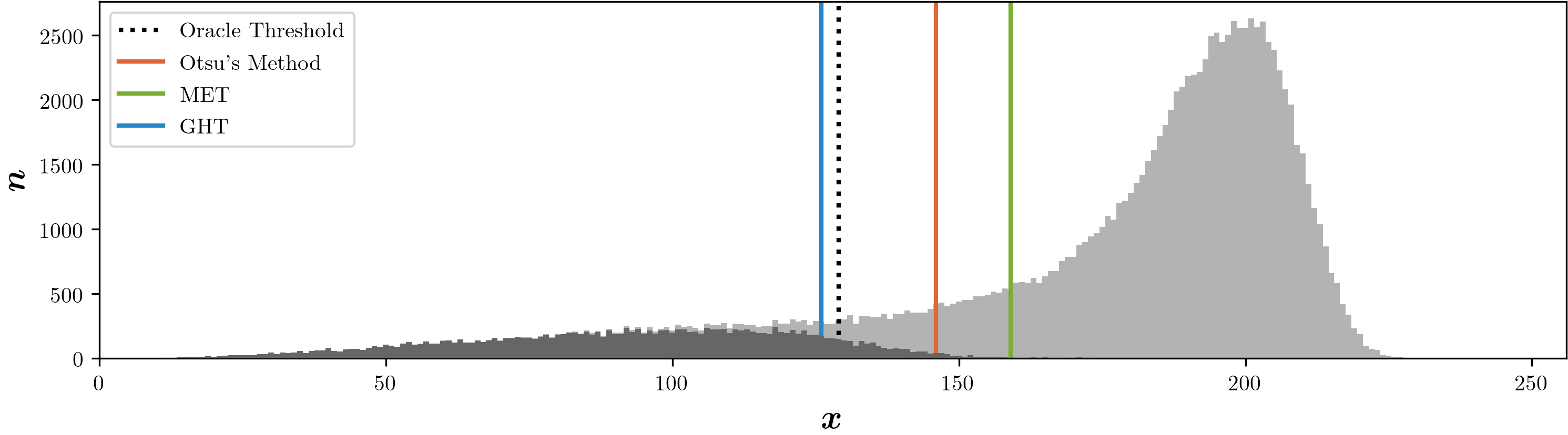}
        \caption{Histogram and thresholds}

    \end{subfigure}
    \caption{
    A visualization of our results on an image from the 2016 Handwritten Document Image Binarization Contest (H-DIBCO) challenge \cite{dibco2016}.
    Given (a) the input image we show (b) the ground-truth binarization provided by the dataset, as well as (c) the binarization according to the lowest error (oracle) global threshold, which represents an upper bound on the performance of any global thresholding-based binarization algorithm.
    The results of Otsu's method, MET, and our GHT are shown in (d-f), where black pixels indicate correct binarization, red pixels indicate false negatives, and blue pixels indicate false positives.
    In (g) we also show a stacked histogram of input pixel intensities (where masked pixels are rendered in a darker color) alongside the output threshold of each algorithm and the oracle global threshold.
    }
    \label{fig:results1}
\end{figure}

\begin{figure}[p]
    \centering 
    \begin{subfigure}[b]{0.325\linewidth}
        \centering
        \includegraphics[width=\linewidth]{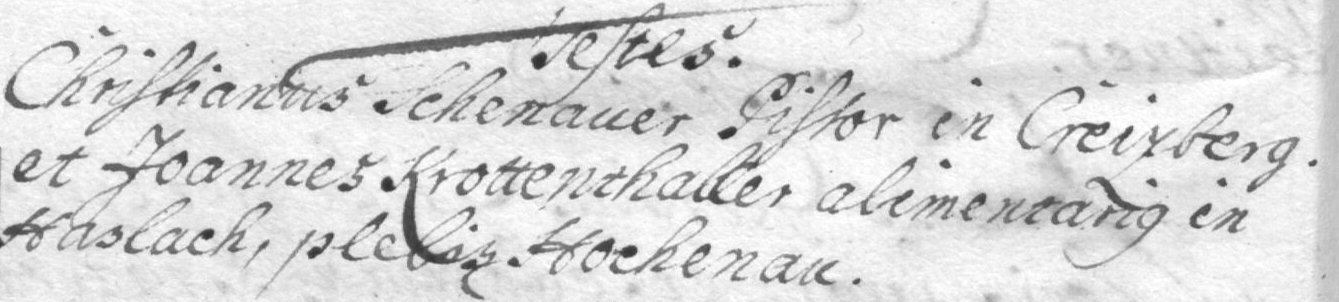}
        \caption{Input Image}

    \end{subfigure}
    \begin{subfigure}[b]{0.325\linewidth}
        \centering
        \frame{\includegraphics[width=\linewidth]{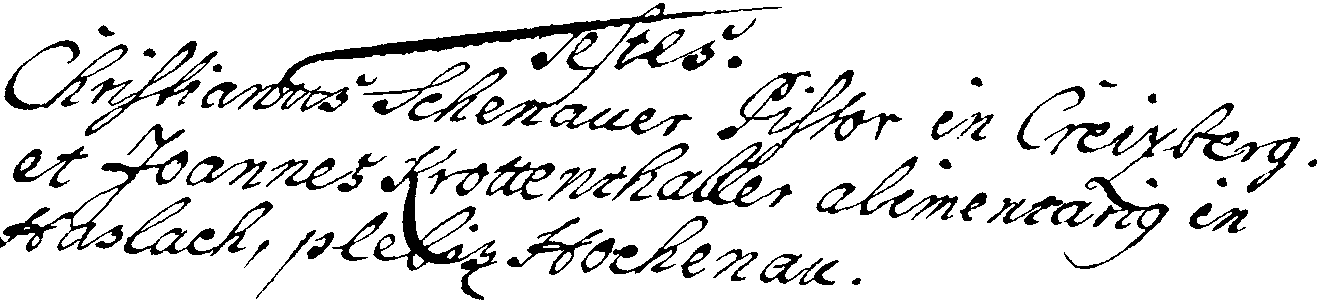}}
        \caption{True Binarization}

    \end{subfigure}
    \begin{subfigure}[b]{0.325\linewidth}
        \centering
        \frame{\includegraphics[width=\linewidth]{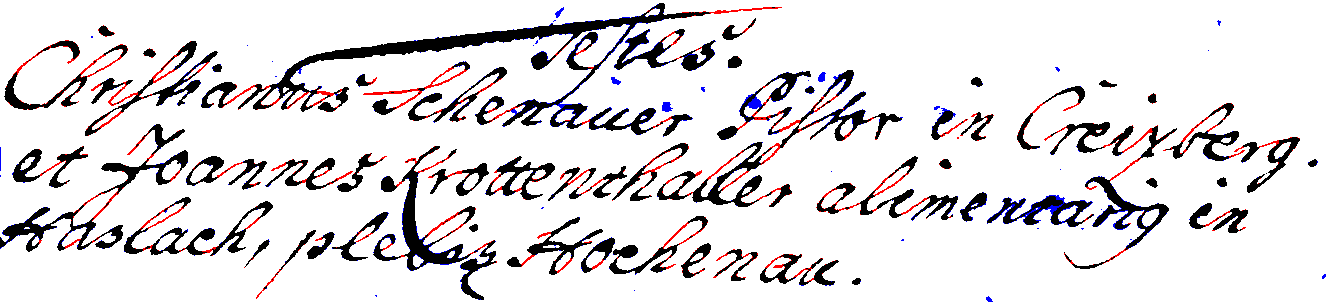}}
        \caption{Oracle Binarization}

    \end{subfigure}
    \begin{subfigure}[b]{0.325\linewidth}
        \centering
        \frame{\includegraphics[width=\linewidth]{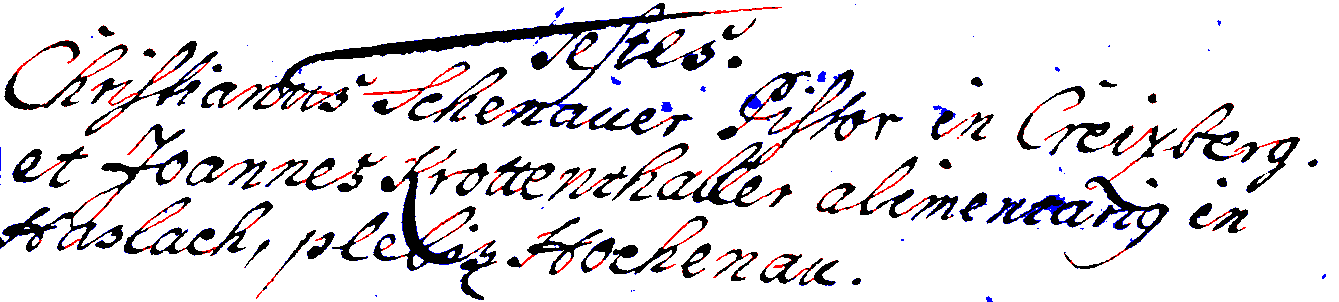}}
        \caption{Otsu's Method}

    \end{subfigure}
    \begin{subfigure}[b]{0.325\linewidth}
        \centering
        \frame{\includegraphics[width=\linewidth]{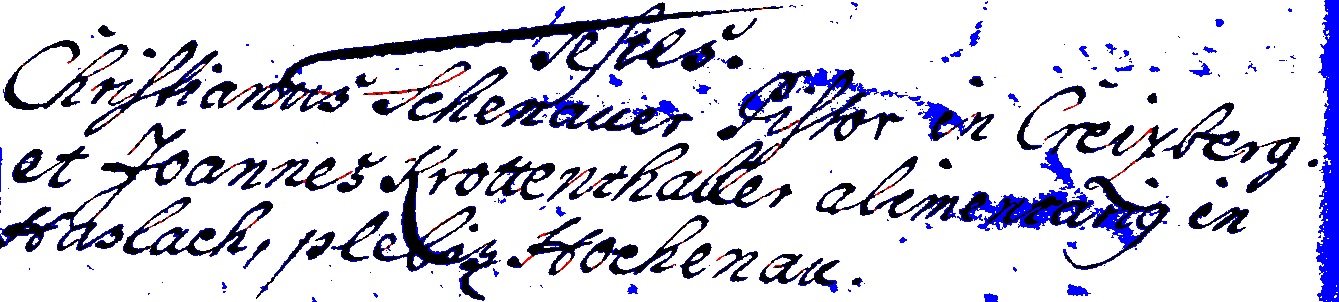}}
        \caption{MET}

    \end{subfigure}
    \begin{subfigure}[b]{0.325\linewidth}
        \centering
        \frame{\includegraphics[width=\linewidth]{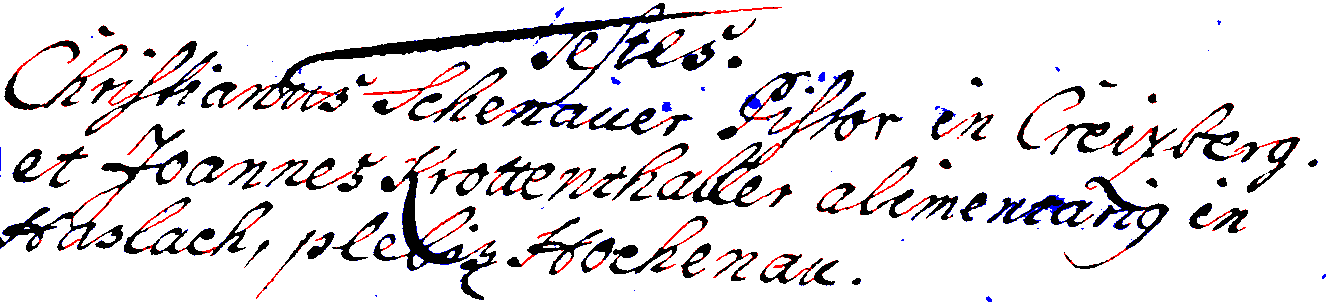}}
        \caption{GHT}

    \end{subfigure}
    \begin{subfigure}[b]{\linewidth}
        \centering
        \includegraphics[width=\linewidth]{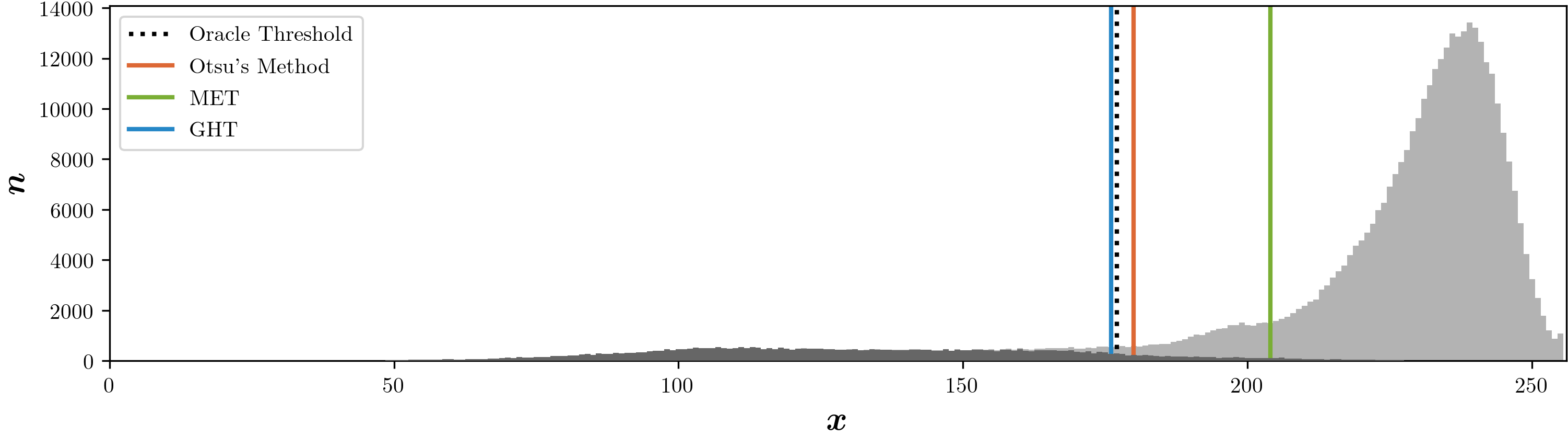}
        \caption{Histogram and thresholds}

    \end{subfigure}
    \caption{
    A visualization of additional results, shown in the same format as Figure~\ref{fig:results1}.
    }
    \label{fig:results4}
\end{figure}

In Table~\ref{table:results} we report GHT's performance against all algorithms evaluated in the challenge~\cite{dibco2016}, using the $F_1$, PSNR, and Distance-Reciprocal Distortion~\cite{lu2004distance} metrics used by the challenge~\cite{dibco2016}.
We see that GHT produces the lowest-error of all entrants to the H-DIBCO 2016 challenge for two of the three metrics used. GHT outperforms Otsu's method and MET by a significant margin, and also outperforms or matches the performance of significantly more complicated techniques that rely on large neural networks with thousands or millions of learned parameters, and that produce an arbitrary per-pixel binary mask instead of GHT's single global threshold. This is despite GHT's simplicity: it requires roughly a dozen lines of code to implement (far fewer if an implementation of MET or Otsu's method is available), requires no training, and has only four parameters that were tuned on a small dataset of only 8 training images. 
See Figures~\ref{fig:results1}-\ref{fig:results4} for visualizations of GHT alongside MET and Otsu's method.

We augment Table~\ref{table:results} with some additional results not present in \cite{dibco2016}. We present an ``Oracle Global Threshold'' algorithm, which shows the performance of an oracle that selects the best-performing (according to $F_1$) global threshold individually for each test image.
We present the special cases of GHT that correspond to MET and Otsu's method, to verify that the special case corresponding to Otsu's method performs identically to our own implementation of Otsu's method and nearly identically to the implementation presented in \cite{dibco2016}.
We present additional ablations of GHT to demonstrate the contribution of each algorithm component: The ``wprctile Case'' model sets $\regstrength$ to an extremely large value and tunes $\regtarget$ on our training set, and performs poorly.
The ``No wprctile'' model sets $\regstrength=0$ and exhibits worse performance than complete GHT.

See Appendix A for reference implementations of GHT, as well as reference implementations of the other algorithms that were demonstrated to be special cases of GHT in Sections~\ref{subsec:met}-\ref{sec:wprctile}.

\section{Conclusion}

We have presented Generalized Histogram Thresholding, a simple, fast, and effective technique for histogram-based image thresholding. GHT includes several classic techniques as special cases (Otsu's method, Minimum Error Thresholding, and weighted percentile thresholding) and thereby serves as a unifying framework for those discrete algorithms, in addition to providing a theoretical grounding for the common practice of varying a histogram's bin width when thresholding.
GHT is exceedingly simple: it can be implemented in just a dozen lines of python (far fewer if an implementation of MET or Otsu's method is available) and has just four tunable hyperparameters.
Because it requires just a single sweep over a histogram of image intensities, GHT is fast to evaluate: its computational complexity is comparable to Otsu's method.
Despite its simplicity and speed (and its inherent limitations as a \emph{global} thresholding algorithm) GHT outperforms or matches the performance of all submitted techniques on the 2016 H-DIBCO image binarization challenge --- including deep neural networks that have been trained to produce arbitrary per-pixel binarizations.

%
%
\bibliographystyle{splncs04}
\bibliography{ref}

\clearpage

\newcommand{\imageid}[0]{9}

\renewcommand{\imageid}[0]{7}
\begin{figure}[p]
    \centering 
    \begin{subfigure}[b]{0.325\linewidth}
        \centering
        \includegraphics[width=\linewidth]{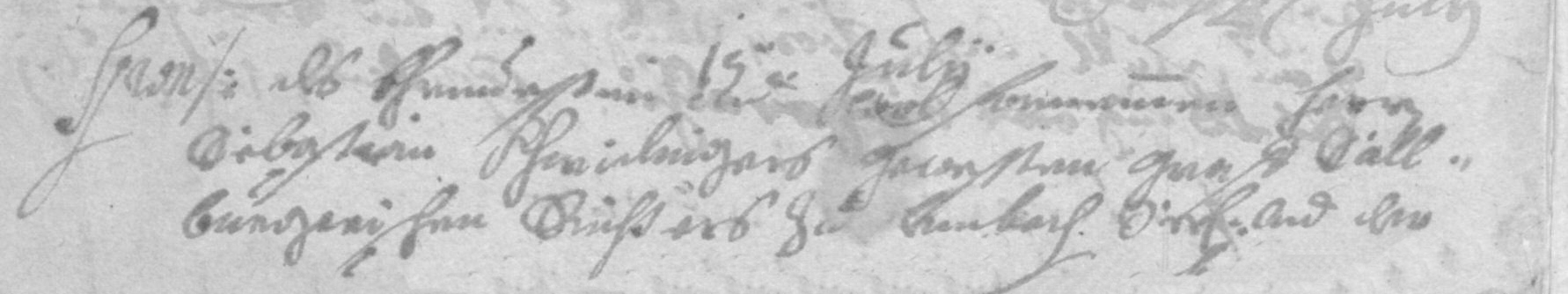}
        \caption{Input Image}

    \end{subfigure}
    \begin{subfigure}[b]{0.325\linewidth}
        \centering
        \frame{\includegraphics[width=\linewidth]{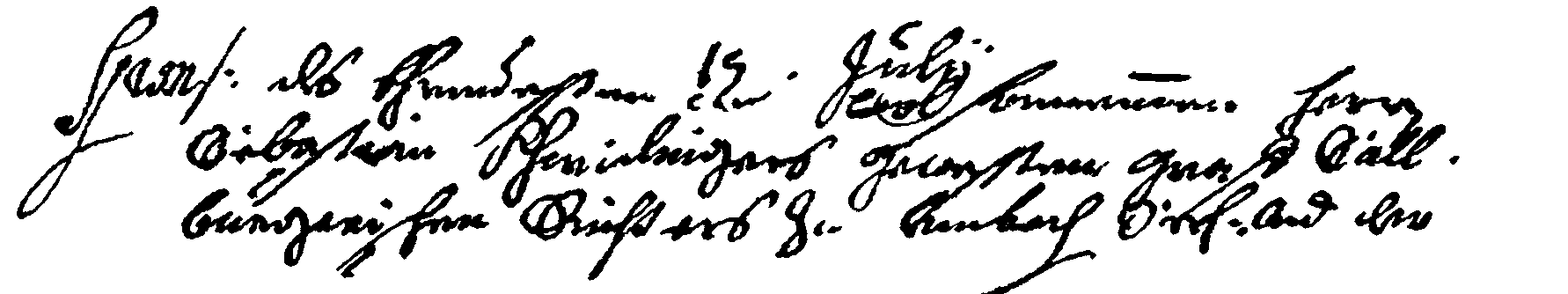}}
        \caption{True Binarization}

    \end{subfigure}
    \begin{subfigure}[b]{0.325\linewidth}
        \centering
        \frame{\includegraphics[width=\linewidth]{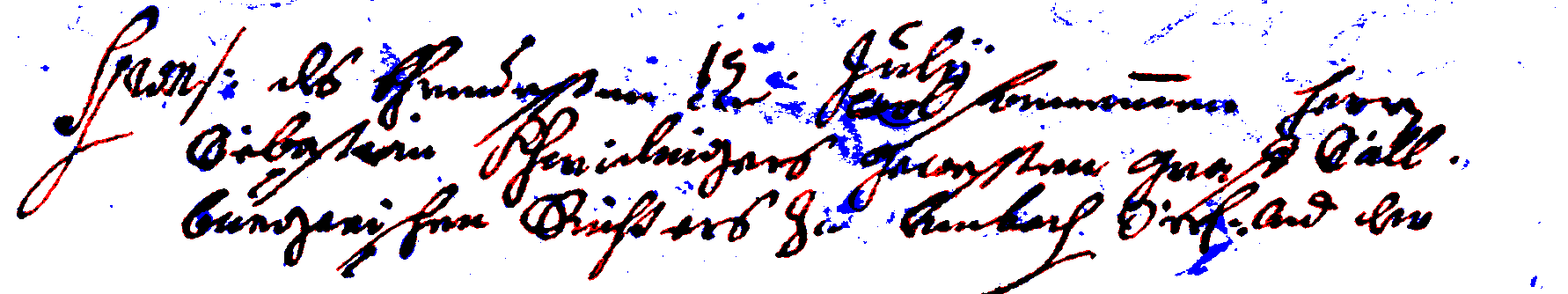}}
        \caption{Oracle Binarization}

    \end{subfigure}
    \begin{subfigure}[b]{0.325\linewidth}
        \centering
        \frame{\includegraphics[width=\linewidth]{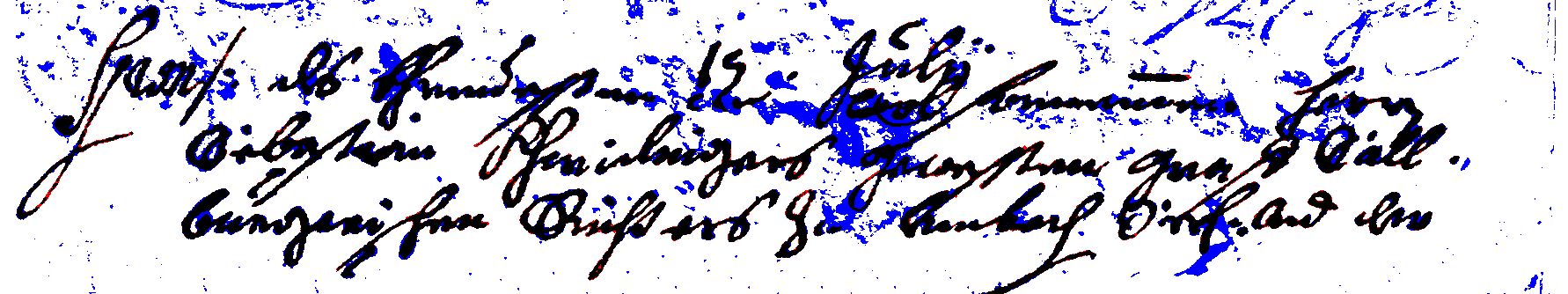}}
        \caption{Otsu's Method}

    \end{subfigure}
    \begin{subfigure}[b]{0.325\linewidth}
        \centering
        \frame{\includegraphics[width=\linewidth]{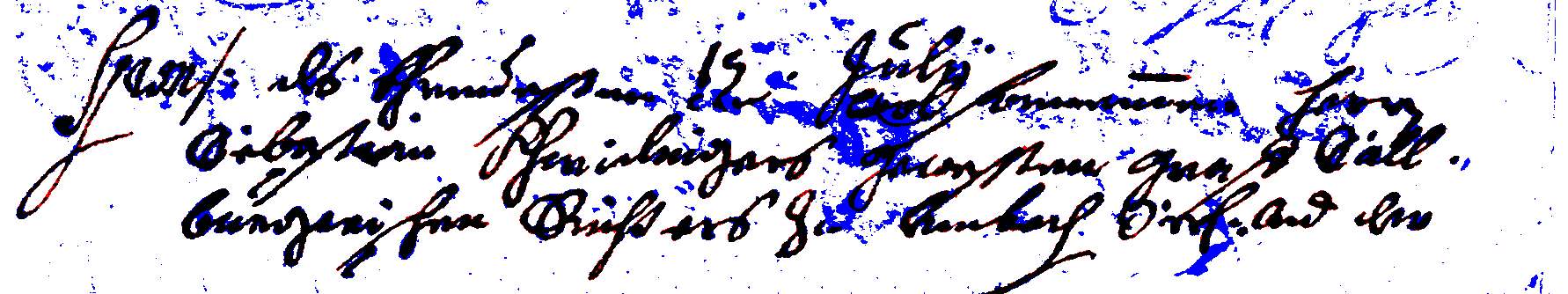}}
        \caption{MET}

    \end{subfigure}
    \begin{subfigure}[b]{0.325\linewidth}
        \centering
        \frame{\includegraphics[width=\linewidth]{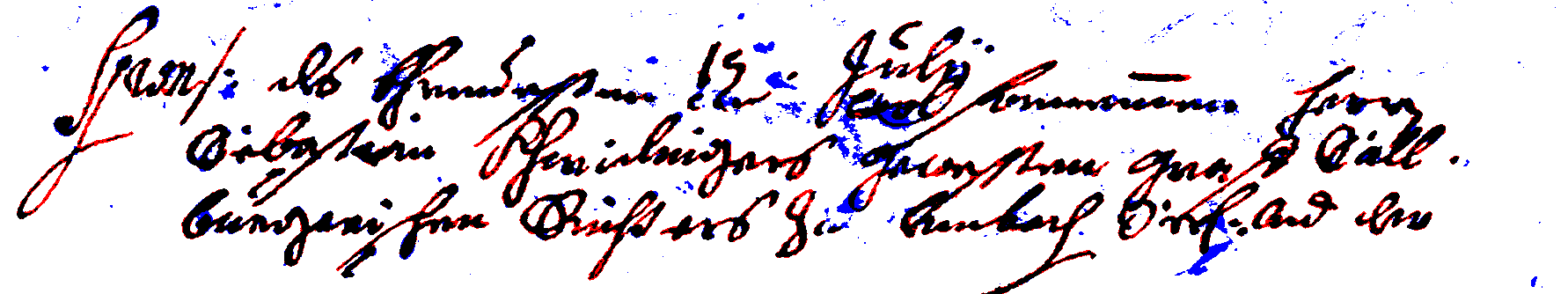}}
        \caption{GHT}

    \end{subfigure}
    \begin{subfigure}[b]{\linewidth}
        \centering
        \includegraphics[width=\linewidth]{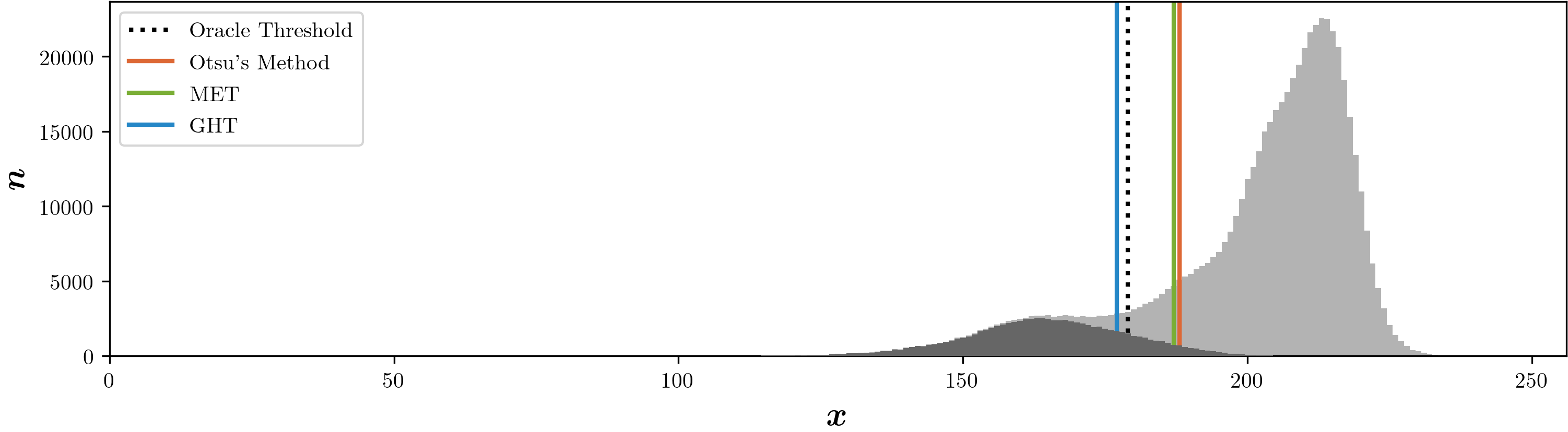}
        \caption{Histogram and thresholds}

    \end{subfigure}
    \caption{
    A visualization of additional results, shown in the same format as Figure~\ref{fig:results1}.
    }
\end{figure}

\renewcommand{\imageid}[0]{8}

\clearpage

\appendix

\section{Reference Code}

In Algorithm~\ref{alg:refcode} we present a reference implementation of GHT, as well as reference implementations of the other algorithms that were demonstrated to be special cases of GHT. In Algorithm~\ref{alg:forloopcode} we present an additional equivalent implementation of GHT that uses an explicit for-loop over splits of the histogram instead of the cumulative sum approach used by the paper. This additional implementation is intended to allow for easier comparisons with similar implementations of MET or Otsu's method, and to allow existing implementations of MET or Otsu's method to be easily generalized into implementations of GHT. We also present a third reference implementation: Algorithm~\ref{alg:probcode} implements GHT in terms of the actual underlying expected complete log-likelihood maximization that reduces to Algorithm~\ref{alg:refcode}. This implementation is highly inefficient, but can be used to verify the correctness of GHT in terms of its Bayesian motivation, and may be useful in deriving further probabilistic extensions.

\lstset{basicstyle=\ttfamily\tiny}

\renewcommand{\tablename}{Algorithm}
\setcounter{table}{0}

\begin{table*}
\begin{lstlisting}
import numpy as np

csum = lambda z: np.cumsum(z)[:-1]
dsum = lambda z: np.cumsum(z[::-1])[-2::-1]
argmax = lambda x, f: np.mean(x[:-1][f == np.max(f)])  # Use the mean for ties.
clip = lambda z: np.maximum(1e-30, z)

def preliminaries(n, x):
  """Some math that is shared across each algorithm."""
  assert np.all(n >= 0)
  x = np.arange(len(n), dtype=n.dtype) if x is None else x
  assert np.all(x[1:] >= x[:-1])
  w0 = clip(csum(n))
  w1 = clip(dsum(n))
  p0 = w0 / (w0 + w1)
  p1 = w1 / (w0 + w1)
  mu0 = csum(n * x) / w0
  mu1 = dsum(n * x) / w1
  d0 = csum(n * x**2) - w0 * mu0**2
  d1 = dsum(n * x**2) - w1 * mu1**2
  return x, w0, w1, p0, p1, mu0, mu1, d0, d1

def Otsu(n, x=None):
  """Otsu's method."""
  x, w0, w1, _, _, mu0, mu1, _, _ = preliminaries(n, x)
  o = w0 * w1 * (mu0 - mu1)**2
  return argmax(x, o), o

def Otsu_equivalent(n, x=None):
  """Equivalent to Otsu's method."""
  x, _, _, _, _, _, _, d0, d1 = preliminaries(n, x)
  o = np.sum(n) * np.sum(n * x**2) - np.sum(n * x)**2 - np.sum(n) * (d0 + d1)
  return argmax(x, o), o

def MET(n, x=None):
  """Minimum Error Thresholding."""
  x, w0, w1, _, _, _, _, d0, d1 = preliminaries(n, x)
  ell = (1 + w0 * np.log(clip(d0 / w0)) + w1 * np.log(clip(d1 / w1))
      - 2 * (w0 * np.log(clip(w0))      + w1 * np.log(clip(w1))))
  return argmax(x, -ell), ell  # argmin()

def wprctile(n, x=None, omega=0.5):
  """Weighted percentile, with weighted median as default."""
  assert omega >= 0 and omega <= 1
  x, _, _, p0, p1, _, _, _, _ = preliminaries(n, x)
  h = -omega * np.log(clip(p0)) - (1. - omega) * np.log(clip(p1))
  return argmax(x, -h), h  # argmin()

def GHT(n, x=None, nu=0, tau=0, kappa=0, omega=0.5):
  """Our generalization of the above algorithms."""
  assert nu >= 0
  assert tau >= 0
  assert kappa >= 0
  assert omega >= 0 and omega <= 1
  x, w0, w1, p0, p1, _, _, d0, d1 = preliminaries(n, x)
  v0 = clip((p0 * nu * tau**2 + d0) / (p0 * nu + w0))
  v1 = clip((p1 * nu * tau**2 + d1) / (p1 * nu + w1))
  f0 = -d0 / v0 - w0 * np.log(v0) + 2 * (w0 + kappa *      omega)  * np.log(w0)
  f1 = -d1 / v1 - w1 * np.log(v1) + 2 * (w1 + kappa * (1 - omega)) * np.log(w1)
  return argmax(x, f0 + f1), f0 + f1
\end{lstlisting}
\caption{Reference Python 3 code for our $\operatorname{GHT}$ algorithm and the baseline algorithms it generalizes: MET,  Otsu's method (here implemented in two equivalent ways), and weighted percentile.
\label{alg:refcode}
}
\end{table*}

\begin{table*}
\begin{lstlisting}
import numpy as np

clip = lambda z: np.maximum(1e-30, z)

def GHT_forloop(n, x=None, nu=0, tau=0, kappa=0, omega=0.5):
  """An implementation of GHT() written using for loops."""
  assert np.all(n >= 0)
  x = np.arange(len(n), dtype=n.dtype) if x is None else x
  assert np.all(x[1:] >= x[:-1])
  assert nu >= 0
  assert tau >= 0
  assert kappa >= 0
  assert omega >= 0 and omega <= 1

  n_sum = np.sum(n)
  nx_sum = np.sum(n * x)
  nxx_sum = np.sum(n * x**2)

  max_score, n_c, nx_c, nxx_c = -np.inf, 0, 0, 0
  for i in range(len(n) - 1):
    n_c += n[i]
    nx_c += n[i] * x[i]
    nxx_c += n[i] * x[i]**2
    w0 = clip(n_c)
    w1 = clip(n_sum - n_c)
    p0 = w0 / n_sum
    p1 = w1 / n_sum
    d0 = np.maximum(0, nxx_c - nx_c**2 / w0)
    d1 = np.maximum(0, (nxx_sum - nxx_c) - (nx_sum - nx_c)**2 / w1)
    v0 = clip((p0 * nu * tau**2 + d0) / (p0 * nu + w0))
    v1 = clip((p1 * nu * tau**2 + d1) / (p1 * nu + w1))
    f0 = -d0 / v0 - w0 * np.log(v0) + 2 * (w0 + kappa *      omega)  * np.log(w0)
    f1 = -d1 / v1 - w1 * np.log(v1) + 2 * (w1 + kappa * (1 - omega)) * np.log(w1)
    score = f0 + f1

    # Argmax where the mean() is used for ties.
    if score > max_score:
      max_score, t_numer, t_denom = score, 0, 0
    if score == max_score:
      t_numer += x[i]
      t_denom += 1
  return t_numer / t_denom
\end{lstlisting}
\caption{Reference Python 3 code for our algorithm GHT written using only for-loops. This is likely less efficient than the implementation provided in Algorithm~\ref{alg:refcode}, and is only provided to demonstrate how this algorithm can be implemented in a single sweep over the histogram (ignoring the calls to \texttt{np.sum()}), and to ease integration into existing implementations of Otsu's method or MET.
\label{alg:forloopcode}
}
\end{table*}

\begin{table*}
\begin{lstlisting}
import numpy as np
from tensorflow_probability import distributions as tfd

def sichi2_var(n, resid, nu, tau):
  """Posterior estimate of variance for a scaled inverse chi-squared."""
  return (nu * tau**2 + np.sum(n * resid**2)) / (nu + np.sum(n))

def GHT_prob(n, x=None, nu=0, tau=0, kappa=0, omega=0.5):
  """An implementation of GHT() using probability distributions."""
  assert np.all(n >= 0)
  x = np.arange(len(n), dtype=n.dtype) if x is None else x
  assert np.all(x[1:] >= x[:-1])
  assert nu >= 0
  assert tau >= 0
  assert kappa >= 0
  assert omega >= 0 and omega <= 1

  n_sum = np.sum(n)
  lls = np.zeros(len(n) - 1)
  for i in range(len(lls)):
    n0, n1 = n[:(i+1)], n[(i+1):]
    x0, x1 = x[:(i+1)], x[(i+1):]
    w0 = clip(np.sum(n0))
    w1 = clip(np.sum(n1))
    p0 = clip(w0 / n_sum)
    p1 = clip(w1 / n_sum)
    mu0 = np.sum(n0 * x0) / w0
    mu1 = np.sum(n1 * x1) / w1
    var0 = sichi2_var(n0, x0 - mu0, p0 * nu, tau)
    var1 = sichi2_var(n1, x1 - mu1, p1 * nu, tau)
    lls[i] = ((np.sum(n0 * (np.log(p0) + tfd.Normal(mu0, np.sqrt(var0)).log_prob(x0)))
             + np.sum(n1 * (np.log(p1) + tfd.Normal(mu1, np.sqrt(var1)).log_prob(x1))))
             + tfd.Beta(kappa * omega + 1, kappa * (1 - omega) + 1).log_prob(np.minimum(p0, 1-1e-15)))
  return np.mean(x[:-1][lls == np.max(lls)]), lls
\end{lstlisting}
\caption{Reference Python 3 and Tensorflow Probability~\cite{dillon2017tensorflow} code for GHT written in terms of probabilities as described in the paper. This is a highly inefficient way to implement this algorithm, but is provided to allow the reader to verify the equivalence between the efficient GHT implementation presented earlier and this Bayesian formulation, and to enable future work that extends this approach.
\label{alg:probcode}
}
\end{table*}

\end{document}